\documentclass[letterpaper, 10 pt, conference]{ieeeconf}  %

\IEEEoverridecommandlockouts                              %

\usepackage{graphicx}
\usepackage{amsmath}
\usepackage{amssymb}
\usepackage{booktabs}
\usepackage{xcolor}
\usepackage{adjustbox}
\usepackage[colorlinks=true,linkcolor=black,citecolor=black,urlcolor=black]{hyperref}
\usepackage[capitalize,noabbrev]{cleveref}
\usepackage{tcolorbox}
\tcbuselibrary{breakable}
\tcbset{promptbox/.style={colback=gray!10,colframe=gray!60,boxrule=0.5pt,arc=0pt,left=4pt,right=4pt,top=4pt,bottom=4pt,breakable}}
\newif\ifjisoonotes
\jisoonotesfalse
\definecolor{jisoocolor}{HTML}{C2185B}

\definecolor{refcolor}{HTML}{0047AB}

\title{
{\Large \bfseries ZeroDex: Zero-Shot Long-Horizon Dexterous Manipulation}\\[-0.1em]
{\Large \bfseries via Multi-View 3D-Grounded VLM Reasoning}
}

\author{Jisoo Kim$^{1}$, Sangwon Baik$^{1}$, Taeksoo Kim$^{1}$, Sungjoo Kim$^{1}$, Junyoung Lee$^{1}$, Mingi Choi$^{1}$,\\
Kisang Park$^{2}$, Sungjoon Choi$^{2}$, and Hanbyul Joo$^{1}$%
\thanks{$^{1}$Seoul National University, Seoul, Republic of Korea.}%
\thanks{$^{2}$Korea University, Seoul, Republic of Korea.}%
}

\begin{document}

\maketitle
\thispagestyle{empty}
\pagestyle{empty}

\begin{abstract}
We present ZeroDex, a zero-shot framework for long-horizon dexterous manipulation that grounds language instructions into executable 3D task plans from calibrated multi-view RGB images. Our framework builds on the grounding ability that vision-language models (VLMs) already exhibit in 2D and lifts their predictions into 3D through multi-view fusion, which extends VLM grounding beyond image-space keypoints to the metric 3D quantities that manipulation actually requires: contact points, placement targets, and tool-motion anchors. Because execution is driven by these 3D groundings rather than a learned policy, the system preserves the VLM's open-world generalization, without collecting demonstrations or post-training for a specific task or embodiment. The VLM decomposes a complex instruction into a sequence of atomic tasks, each executed by a primitive from a small reusable library that covers manipulation tasks including pick-and-place and tool-use: for tool-use, we retrieve an object-centric atomic action corresponding to the inferred skill category and align its stored 6D tool trajectory to the grounded keypoints; for dexterous execution, we expand the lifted grasp keypoint into a task-conditioned grasp affordance region and generate feasible grasp-motion pairs with an arm-hand motion generator. Real-world experiments show improved 3D grounding accuracy and higher task success than zero-shot visual-prompting and constraint-based baselines, both of which ground from a single RGB-D view, and our demonstration-free system outperforms VLA policies fine-tuned on 100 task demonstrations. We further demonstrate long-horizon manipulation, where execution verification and replanning recover from intermediate failures, and show that the system generalizes to unseen objects and tool-use tasks in novel scenes.

\end{abstract}


\section{Introduction}

A long-standing goal in robotics is to build general-purpose systems that perform long-horizon manipulation from high-level language instructions. Beyond recognizing objects, such systems must ground instructions in task-relevant 3D geometry: where to place an object, which part to contact, and how to orient and move a tool during execution. This requirement is especially stringent for dexterous hands, where small 3D grounding errors cause unstable grasps, collisions, inverse-kinematics failures, or contact on the wrong functional region of a tool.

The dominant approach learns this behavior end-to-end. Recent vision-language-action (VLA) models make strong progress toward general-purpose manipulation by predicting actions directly from large-scale robot data~\cite{brohan2022rt,zitkovich2023rt,kim2024openvla,team2024octo,black2024pi_0}. However, achieving reliable performance on diverse manipulation tasks requires extensive data collection, task-specific adaptation, or environment and embodiment specific fine-tuning, which makes them hard to scale across the diversity of objects, tools, and spatial configurations found in open-ended settings. Human demonstration retargeting offers another route, but the embodiment gap between human and robotic hands produces physically infeasible contacts or unstable grasps, and often demands additional refinement or reinforcement learning before deployment.
In this paper, we explore an alternative strategy that builds a modular interface between vision-language reasoning and motion control. Instead of learning a single end-to-end policy, we decouple semantic reasoning, handled by a vision-language model (VLM), from physical execution, handled by motion primitives or controllers. The key observation is that modern VLMs already answer, zero-shot, most of the sub-questions that manipulation requires: what to grasp, where to grasp it as a functional affordance, how to move it, and in what order to carry out the steps of a task. Given this competence, we do not need to train a new policy. Once the VLM produces a plan, we decompose a complex task into a sequence of simple atomic tasks such as pick and move, and we execute each atomic motion with a reliable primitive controller. Because most manipulation tasks can be expressed as compositions of such atomic tasks, a small library of primitives driven by VLM reasoning covers a broad range of tasks. We find that VLMs produce accurate plans and decompositions across a wide range of tasks, as demonstrated in our evaluations.

\begin{figure*}[t]
    \centering
    \includegraphics[width=\textwidth]{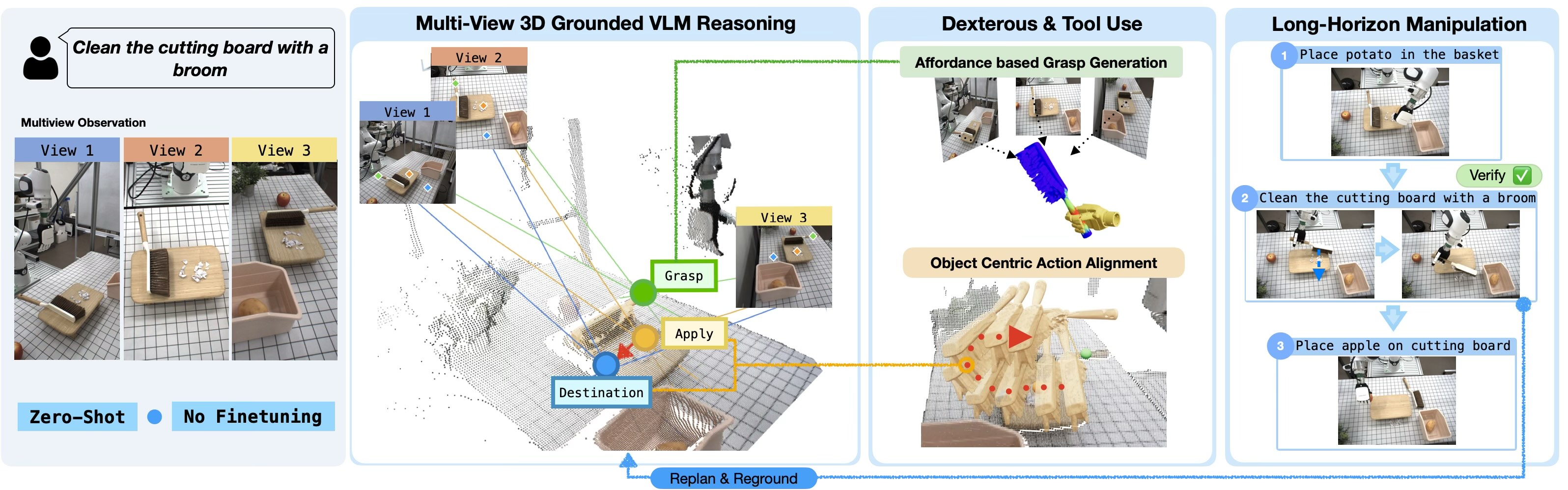}
    \caption{\textbf{ZeroDex.} Given a language instruction and calibrated multi-view observations, our framework uses multi-view VLM grounding with robust triangulation to infer task-relevant 3D groundings, generates affordance-aware dexterous grasps, and executes pick-and-place or tool-use plans through reusable action primitives.}
    \label{fig:overview}
\end{figure*}

This modular design, however, hinges on a geometric requirement that 2D reasoning cannot satisfy on its own, as the relevant quantities are inherently three-dimensional. Where to grasp must be specified in 3D rather than 2D, and more importantly, how to move the end effector and where to move an object are 3D trajectory quantities. A single view rarely carries enough geometric information to reason about these 3D trajectories reliably, so attaching a VLM to a single image is limited for this purpose.
Our central idea is to fuse VLM grounding across multiple views. Multi-view fusion lifts view-dependent 2D predictions into consistent 3D quantities, and it extends VLM-based grounding to a far wider set of problems than single-view reasoning can reach, including 3D contact points, placement targets, and tool-motion anchors. Within this framework we triangulate VLM groundings across views to recover 3D positions and directions. Since VLM 2D groundings are semantic and view-dependent, we take the 3D candidate with the widest cross-view agreement rather than averaging, which keeps grounding reliable under occlusion and partial visibility.

The lifted keypoints are by themselves sufficient for most pick-and-place tasks, where a grasp point, a transport waypoint, and a placement target fully specify the motion. Tasks such as pouring or sweeping instead depend on how the tool moves relative to the target over time, a 6D trajectory that a few keypoints cannot express; we obtain it by reusing short, environment-independent object trajectories and aligning them to the grounded geometry. We represent these tool-use behaviors as a \emph{Bag of Atomic Actions}, a library of short 6D object trajectories indexed by interaction type. For a new scene, the appropriate primitive is retrieved and aligned to the grounded task geometry. To support dexterous-hand execution, we apply the same multi-view grounding to estimate functional contact regions, generate candidate grasps on those regions, and filter them by inverse-kinematics and collision feasibility over the full tool-use trajectory. For long-horizon tasks, the VLM verifies the outcome of each executed subtask and maintains a textual memory of the overall scenario and of what has been executed so far; together these let the system re-ground, retry, or replan the remaining subtasks after execution failures or scene changes.

We instantiate these ideas in ZeroDex, a single zero-shot system, and evaluate it on a diverse set of real-world dexterous manipulation tasks. Our main contributions are as follows. First, we present ZeroDex, a framework for long-horizon dexterous manipulation that couples VLM-based task planning and multi-view 3D grounding with a library of reusable atomic action primitives, enabling zero-shot execution in unseen scenes. Second, we propose a multi-view VLM grounding approach that fuses view-dependent 2D predictions into reliable 3D quantities through consensus-based cross-view triangulation, remaining robust under occlusion and partial visibility. Third, we show that the same multi-view grounding serves both halves of execution: the lifted interaction and terminal keypoints provide the correspondences that align a stored 6D tool trajectory to the current scene, while the lifted grasp keypoint is expanded into a functional 3D affordance region for grasp generation, with candidates filtered by trajectory-level inverse-kinematics and collision feasibility. Finally, we validate the complete system on a diverse set of real-world manipulation tasks, where multi-view fusion yields over $3\times$ more repeatable 3D grounding than the single-view RGB-D lifting of MOKA~\cite{liu2403moka}, and higher success than zero-shot visual-prompting and constraint-based baselines, while our demonstration-free system outperforms VLA policies fine-tuned on 100 task-specific demonstrations.

\section{Related Work}
\label{sec:citations}

\noindent \textbf{Vision-Language-Action Models for Manipulation}
Vision-language-action (VLA) models map images and instructions to robot actions, often by adapting
vision-language backbones~\cite{beyer2024paligemma,chen2024internvl,bai2025qwen3} with large
robot datasets~\cite{ebert2021bridge,walke2023bridgedata,o2024open,khazatsky2024droid,bu2025agibot}.
Beyond foundational robot policies~\cite{brohan2022rt,zitkovich2023rt,team2024octo,kim2024openvla,black2024pi_0},
recent work improves action modeling~\cite{black2024pi_0,wen2025dexvla}, fine-tuning
efficiency~\cite{kim2025fine}, spatial grounding~\cite{qu2025spatialvla,li2024cogact}, open-world
generalization~\cite{team2025gemini}, and dexterous or humanoid
embodiments~\cite{bjorck2025gr00t,wen2025dexvla,zhong2026dexgraspvla,luo2025being,kim2026rldx}.
Despite strong in-distribution performance, such policies, including recent variants that add reasoning or 3D structure inside the policy~\cite{zawalski2024ecot,fan2026peafowl}, rely on large robot data and still require
additional data or adaptation to cover new objects, tools, and embodiments. Retargeting human
demonstrations~\cite{qin2022dexmv,sivakumar2022robotic,wang2024dexcap} reduces collection cost, but
the embodiment gap can yield infeasible contacts that require further refinement. We instead act
zero-shot in unseen scenes, without task-specific demonstrations or weight updates.

\noindent \textbf{Zero-Shot Manipulation with Foundation Models}
Another line drives manipulation directly with pretrained foundation models, without task-specific
training, and differs mainly in the intermediate representation it extracts. Code-generation methods
prompt large language models to produce high-level plans or executable control code over perception and
motion APIs~\cite{liang2023code,singh2023progprompt,huang2023voxposer}. Visual-prompting methods query
vision-language models to mark the task on images as keypoints, visual marks, or
affordances~\cite{liu2403moka,huang2024rekep,nasiriany2024pivot,yuan2024robopoint, duan2024manipulate}. Zero-shot orchestrators~\cite{shi2025maestro,galanti2026pigey} use a VLM to sequence and verify skills, yet delegate execution to learned policies and typically ground from a single view. These
methods share our zero-shot, training-free motivation, but many still rely on image-space or sparse intermediate representations that are brittle for the task-relevant 3D geometry required for dexterous manipulation, including contact points, placement targets, and tool trajectories. This gap motivates grounding tasks directly in 3D.

\noindent \textbf{3D Grounding and Dexterous Execution}
Lifting 2D observations to 3D is classically studied in multi-view stereo~\cite{furukawa2015multi};
for manipulation, recent work gives VLMs spatial reasoning~\cite{chen2024spatialvlm,song2025robospatial},
aggregates views into 3D groundings~\cite{xu2024vlm}, or casts VLM reasoning as 3D
constraints~\cite{huang2023voxposer,huang2024rekep,pan2025omnimanip}, often aided by depth or pose
foundation models~\cite{wen2025foundationstereo,wen2024foundationpose}. For execution, prior work
transfers object-centric skills or functional correspondences~\cite{vlmpose,tang2025mimicfunc},
synthesizes dexterous grasps via samplers, generative models, or affordance
conditioning~\cite{wang2022dexgraspnet,chen2025bodex,chen2025dexonomy,nasiriany2411rt},
and uses language or VLM feedback for verification and replanning~\cite{huang2022inner}. Unlike prior systems that treat 3D grounding, tool motion, and dexterous grasping as separate components, we fuse multi-view VLM groundings into explicit metric 3D quantities, align object-centric 6D tool trajectories to them, and verify collision and kinematic feasibility before execution~\cite{curobo_v2, park2026natural}, all without training.

\section{Method}
\label{sec:method}

Given calibrated multi-view RGB images $\mathcal{I}=\{\mathbf{I}_v\}_{v=1}^{M}$ and a subtask $\mathbf{l}$ (e.g., ``Pour from a kettle''), ZeroDex returns an arm--hand execution trajectory $\mathcal{T}_{\mathrm{robot}}$ and its physical feasibility $\eta \in \{0,1\}$ :
\begin{equation}
    (\mathcal{T}_{\mathrm{robot}},\eta)
    =\mathcal{Z}(\mathcal{I}, \mathbf{l}).
    \label{eq:zerodex_executor}
\end{equation}
Our central representation is a desired object trajectory $\mathcal{T}_{\mathrm{obj}}$ specifying what the manipulated object should do (\cref{sec:method:trajectory}). Its spatial constraints come from reference-view primitives $\mathcal{Q}_r$, which associate action types with relevant image locations (\cref{sec:method:grounding}), and their metric world-frame keypoints $\mathcal{X}$ obtained through multi-view lifting (\cref{sec:method:uplifting}). Grasp and motion generation determines how the arm and hand can perform this object trajectory (\cref{sec:method:grasp}); execution proceeds only if $\eta=1$. The same executor supports long-horizon tasks through planning and execution verification (\cref{sec:method:retry}). \cref{fig:overview} shows our overall pipeline.

\subsection{Reference-View Semantic Grounding}
\label{sec:method:grounding}

We represent a subtask as an ordered sequence of action primitives, each specifying an action type and the image locations needed to perform it. Semantic descriptions identify what these locations refer to, providing correspondence cues for subsequent 3D lifting. In a selected reference view $\mathbf{I}_r$, this representation is
\begin{equation}
    \begin{aligned}
        \mathcal{Q}_r
        &=\{(m^t,\mathcal{P}_r^t)\}_{t=1}^{T}, \\
        \mathcal{P}_r^t
        &=\{(\mathbf{p}_r^{t,j},\mathbf{d}^{t,j})\}_{j=1}^{N_t}.
    \end{aligned}
    \label{eq:planning}
\end{equation}
Here, $T$ is the number of primitives, $m^t\in\{\it{grasp},\allowbreak\it{apply\_action},\allowbreak\it{waypoint},\allowbreak\it{release},\allowbreak\it{hold}\}$ is the primitive type, and $\mathcal{P}_r^t$ contains $N_t$ 2D keypoints $\mathbf{p}_r^{t,j} \in \mathbb{R}^2$ paired with text descriptions $\mathbf{d}^{t,j}$.

Pick-and-place uses $\it{grasp}$, $\it{waypoint}$, and $\it{release}$ to specify a grasp point, a transport waypoint, and a placement location. Tool use employs $\it{grasp}$, $\it{apply\_action}$, and a terminal $\it{release}$ or $\it{hold}$. Its grasp step identifies both a grasp point and a functional tip on the tool (e.g., a broom head or kettle spout); the action step identifies an interaction point on the target; and the terminal step specifies the final location and whether to release or maintain the grasp.

To construct this representation, the VLM $\Phi$ selects the reference view and identifies the task entities:
\begin{equation}
    (r,z,g_z)=\Phi(\mathcal{I},\mathbf{l}),
    \label{eq:task-level_grounding}
\end{equation}
where $r\in\{1,\dots,M\}$ is the view index and $z\in\{\it{pick},\it{tool}\}$ selects the manipulation mode. The grounding tuple is $g_{\mathrm{pick}}=(O_{\mathrm{tar}},\mathbf{p}_{\mathrm{dst}})$ for pick-and-place and $g_{\mathrm{tool}}=(O_{\mathrm{tool}},c,O_{\mathrm{tar}},\mathbf{p}_{\mathrm{dst}})$ for tool use. Here, $O_{\mathrm{tar}}$ is the target object, $\mathbf{p}_{\mathrm{dst}}$ is its destination pixel in $\mathbf{I}_r$, and $O_{\mathrm{tool}}$ and $c$ specify the tool object and skill category. If the target remains stationary, $\mathbf{p}_{\mathrm{dst}}=\mathtt{NULL}$. A planning prompt $\mathbf{l}'$ combines $\mathbf{l}$ and $g_z$ to produce
\begin{equation}
    \mathcal{Q}_r=\Phi(\mathbf{I}_r, \mathbf{\mathbf{l}'}).
    \label{eq:task-level_grounding_2}
\end{equation}
We denote the hand-manipulated object by $O_{\mathrm{man}} \in \{O_{\mathrm{tar}}, O_{\mathrm{tool}}\}$: it is $O_{\mathrm{tar}}$ for pick-and-place and $O_{\mathrm{tool}}$ for tool use.

\subsection{Multi-View Fusion-Based 3D Lifting}
\label{sec:method:uplifting}

The semantic plan identifies relevant image locations but does not determine their metric depth. We recover world-frame 3D keypoints $\mathcal{X}=\{\mathbf{X}_{\star}^{t,j}\}_{t,j}$ by seeking geometric agreement across calibrated views: a candidate 3D point should reproject near the independently grounded 2D locations in multiple images. We omit the per-keypoint indices $(t,j)$ below and let $\pi_v$ denote calibrated world-to-image projection.

A grounding prompt $\mathbf{l}''$ formed from $\mathbf{l}$, $m^t$, and $\mathbf{d}^{t,j}$ yields predictions $\bar{\mathbf{p}}_v=\Phi(\mathbf{I}_v,\mathbf{l}'')$ in each view. Triangulating each view pair $a<b$ gives a world-frame 3D keypoint candidate $\mathbf{X}_{a,b}=\operatorname{Triangulate}(\bar{\mathbf{p}}_a,\bar{\mathbf{p}}_b)$. Its RANSAC~\cite{ransac}-style consensus score counts the views whose reprojection errors fall within a pixel threshold $\epsilon_{\mathrm{tri}}$:
\begin{equation}
    S_{\mathrm{tri}}(\mathbf{X})
    =\sum_{v=1}^{M}\mathbf{1}
      (\|\pi_v(\mathbf{X})-\bar{\mathbf{p}}_v\|_2\leq\epsilon_{\mathrm{tri}}),
    \label{eq:triangulation_score}
\end{equation}
where $\mathbf{1}(\cdot)$ is an indicator function.
The candidate with the largest support provides the lifted keypoint:
\begin{equation}
    \begin{aligned}
        (a^*,b^*)&=\arg\max_{a<b}S_{\mathrm{tri}}(\mathbf{X}_{a,b}), \\
        \mathbf{X}_{\star}&=\mathbf{X}_{a^*,b^*}.
    \end{aligned}
    \label{eq:final_uplifting_selection}
\end{equation}
Calibrated camera extrinsics are incorporated into triangulation, so all lifted keypoints share the world frame.

\subsection{Object Trajectory Construction}
\label{sec:method:trajectory}

The lifted keypoints specify where manipulation should occur; the desired 6D trajectory $\mathcal{T}_{\mathrm{obj}}$ specifies how $O_{\mathrm{man}}$ moves between these locations. For pick-and-place, it connects the current object pose to the placement location through the transport waypoint while preserving the object's initial orientation. Tool use additionally requires a motion pattern describing how the tool acts on the target. Both use placement or terminal locations refined before trajectory construction.

\begin{figure}[t]
    \centering
    \includegraphics[width=\linewidth]{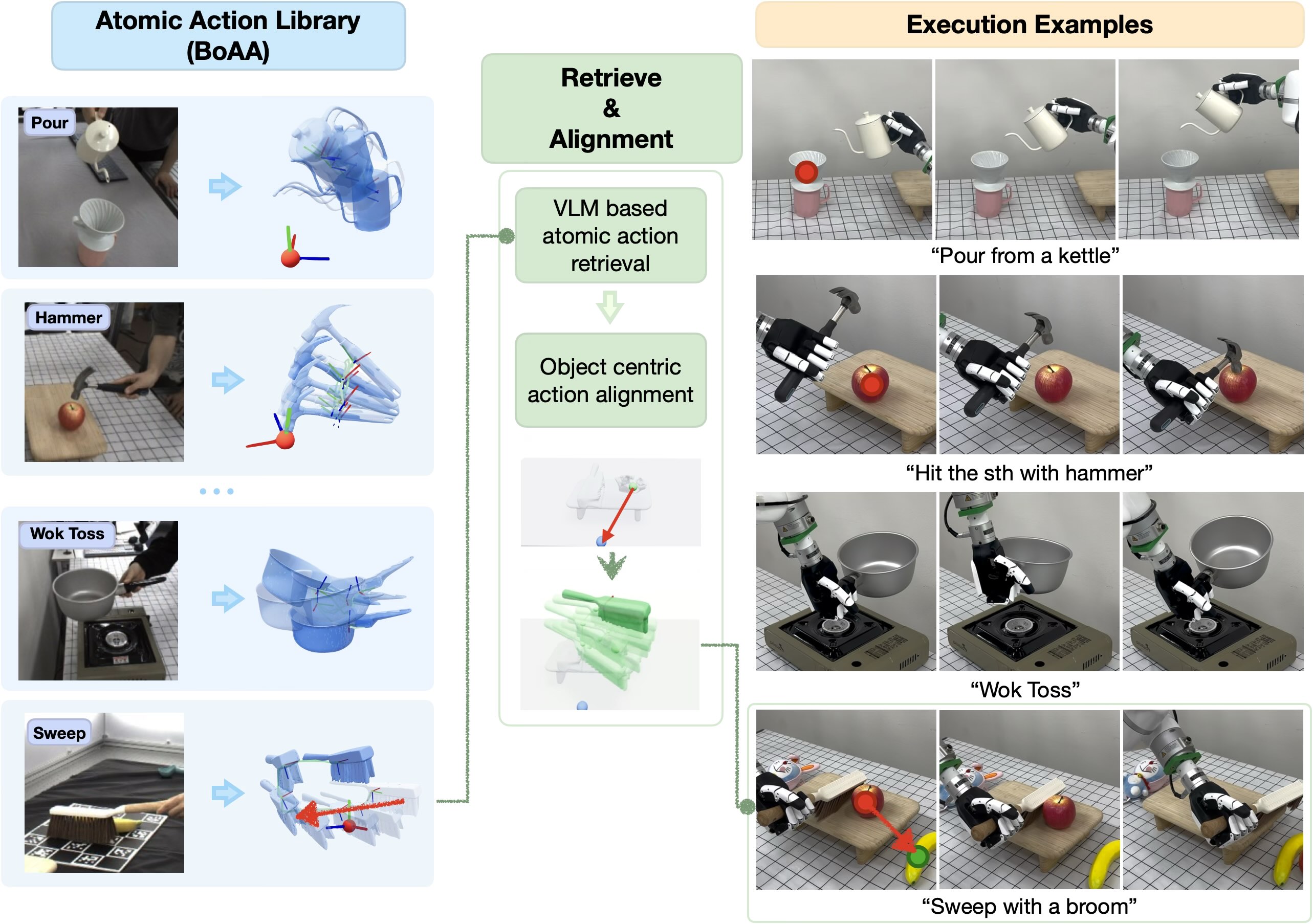}
    \caption{\textbf{Bag of Atomic Actions.} Examples of object-centric atomic actions executed on the real robot.}
    \label{fig:boaa_realtest}
\end{figure}

\paragraph{Tool use via atomic action alignment.}
Our \emph{Bag of Atomic Actions} supplies reusable object-centric motion patterns for skills such as pouring and sweeping. Each entry $\mathcal{A}=(c,\mathcal{T},\mathbf{X}^{\mathrm{s}},\mathbf{X}^{\mathrm{e}})$ stores a skill category $c$, a sequence $\mathcal{T}=(T_i)_{i=0}^{N_a}$ of tool poses $T_i\in SE(3)$, and 3D start/end anchors $\mathbf{X}^{\mathrm{s}},\mathbf{X}^{\mathrm{e}}\in\mathbb{R}^3$. We build the library offline from recorded demonstrations and trajectories generated using VLMPose~\cite{vlmpose} (\cref{fig:boaa_realtest}).

At test time, we retrieve an entry matching $c$. The stored start anchor $\mathbf{X}^{\mathrm{s}}$ corresponds to the lifted $\it{apply\_action}$ point $\mathbf{X}^{\mathrm{app}}$, and the end anchor $\mathbf{X}^{\mathrm{e}}$ to the refined $\it{release}$ or $\it{hold}$ location $\mathbf{X}^{\mathrm{term}}$. A trajectory alignment operator $f_{\mathrm{align}}$ adapts the stored motion using these correspondences, with the alignment rotation restricted to a predefined axis:
\begin{equation}
    \mathcal{T}_{\mathrm{obj}}
    = f_{\mathrm{align}}\bigl(
        \mathcal{T};
        (\mathbf{X}^{\mathrm{s}},\mathbf{X}^{\mathrm{e}}),
        (\mathbf{X}^{\mathrm{app}},\mathbf{X}^{\mathrm{term}})
      \bigr).
    \label{eq:aligned_trajectory}
\end{equation}

\paragraph{Collision-aware endpoint refinement.}
Placement and terminal keypoints must leave room for the manipulated object at its final pose. For a lifted location $\mathbf{X}^{\mathrm{loc}} \in \mathbb{R}^3$, we form a local grid of candidate positions $\mathcal{G}(\mathbf{X}^{\mathrm{loc}})$. With orientation fixed, we evaluate the object's environment penetration depth $\phi(\mathbf{X}') \in \mathbb{R}$ at each candidate $\mathbf{X}' \in \mathcal{G}(\mathbf{X}^{\mathrm{loc}})$ and select the nearest position with zero penetration:
\begin{equation}
    \mathbf{X}^{{\mathrm{loc}}}_*
    =\underset{\substack{
        \mathbf{X}'\in\mathcal{G}(\mathbf{X}^{\mathrm{loc}})\\
        \phi(\mathbf{X}')=0
    }}{\arg\min}
    \|\mathbf{X}'-\mathbf{X}^{\mathrm{loc}}\|_2.
    \label{eq:position_refinement}
\end{equation}
If no valid position exists, grounding fails. This refinement checks the object at the endpoint; feasibility along the robot's motion is addressed in \cref{sec:method:grasp}.

\subsection{Dexterous Grasp and Motion Generation}
\label{sec:method:grasp}

\begin{figure}[t]
    \centering
    \includegraphics[width=\linewidth]{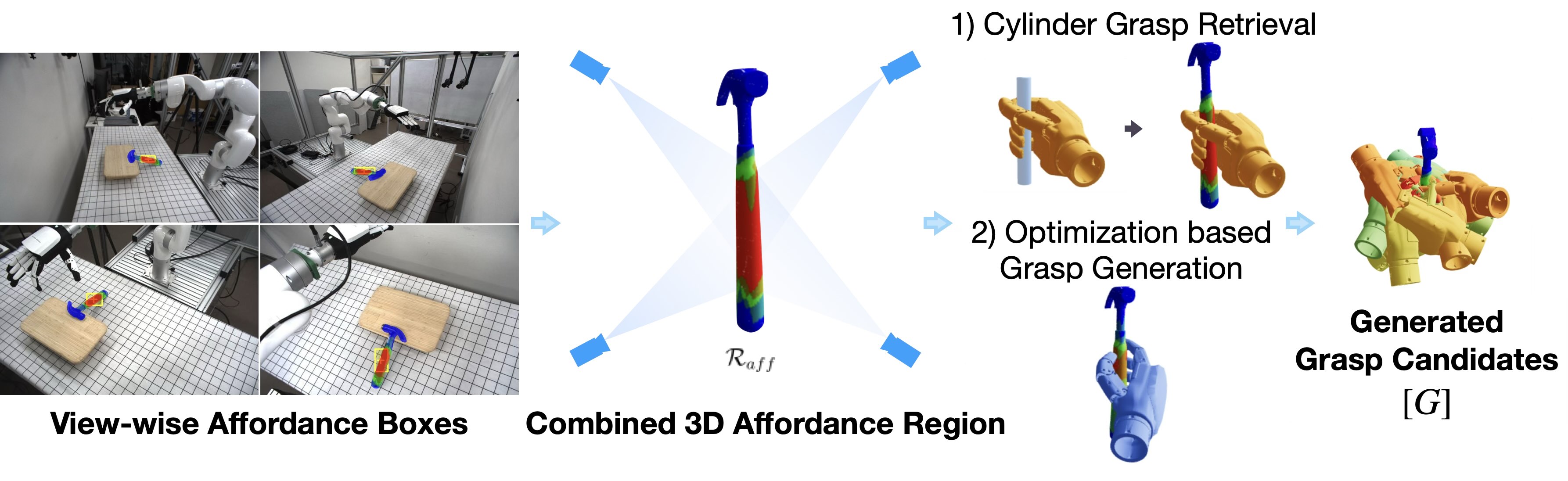}
    \caption{Affordance grounding and grasp generation. Multi-view affordance bounding boxes are combined to identify a 3D affordance region.}
    \label{fig:supp_affordance}
\end{figure}

Executing $\mathcal{T}_{\mathrm{obj}}$ requires a stable grasp and a feasible arm--hand trajectory. The lifted grasp keypoint $\mathbf{X}^{\mathrm{grasp}}\in\mathcal{X}$ identifies a task-relevant object part; we use it to locate a surface region on which the hand can establish contact, then test whether grasps from this region permit the desired motion.

In each view, an affordance prompt $\mathbf{l}'''$ combines $\mathbf{l}$, the projection $\pi_v(\mathbf{X}^{\mathrm{grasp}})$, and its semantic description. The VLM predicts a graspable box $\mathbf{B}_v=\Phi(\mathbf{I}_v,\mathbf{l}''') \in \mathbb{R}^{2\times2}$ on $O_{\mathrm{man}}$. For each world-frame vertex $\mathbf{q}_i \in \mathbb{R}^3$ of the object mesh of $O_{\mathrm{man}}$, we compute the fraction of views whose boxes contain its projection and retain vertices above a threshold $\tau_{\mathrm{aff}}$:
\begin{equation}
    \begin{aligned}
        s(\mathbf{q}_i)&=\frac{1}{M}\sum_{v=1}^{M}
            \mathbf{1}(\pi_v(\mathbf{q}_i)\in \mathbf{B}_v), \\
        \mathcal{R}_{\mathrm{aff}}
        &=\{\mathbf{q}_i\mid s(\mathbf{q}_i)\geq\tau_{\mathrm{aff}}\}.
    \end{aligned}
    \label{eq:affordance_region}
\end{equation}
We generate grasps in $\mathcal{R}_{\mathrm{aff}}$ using cylindrical templates for handle-like affordances and an optimization-based generator~\cite{chen2025bodex} for general geometries. Physics simulation tests their stability under external forces along, and torques about, each coordinate axis. Stable candidates form the grasp candidate set $G$.

For each grasp $\gamma\in G$, an arm--hand motion generator~\cite{curobo_v2,park2026natural} $f_{\mathrm{motion}}$ seeks a robot trajectory that carries $O_{\mathrm{man}}$ along $\mathcal{T}_{\mathrm{obj}}$ while satisfying kinematic and collision constraints throughout the motion:
\begin{equation}
    (\mathcal{T}_{\mathrm{robot}},\eta)
    =f_{\mathrm{motion}}(\mathcal{T}_{\mathrm{obj}},\gamma).
    \label{eq:motion_generation}
\end{equation}
We execute a trajectory with $\eta=1$, following the planned grasp and terminal release/hold behavior. If no grasp yields a feasible trajectory, the module reports failure. Tool use follows the same procedure, with affordances computed on the tool mesh.

\subsection{Extension to Long-Horizon Tasks}
\label{sec:method:retry}

\begin{figure}
\centering
\includegraphics[width=1.0\linewidth]{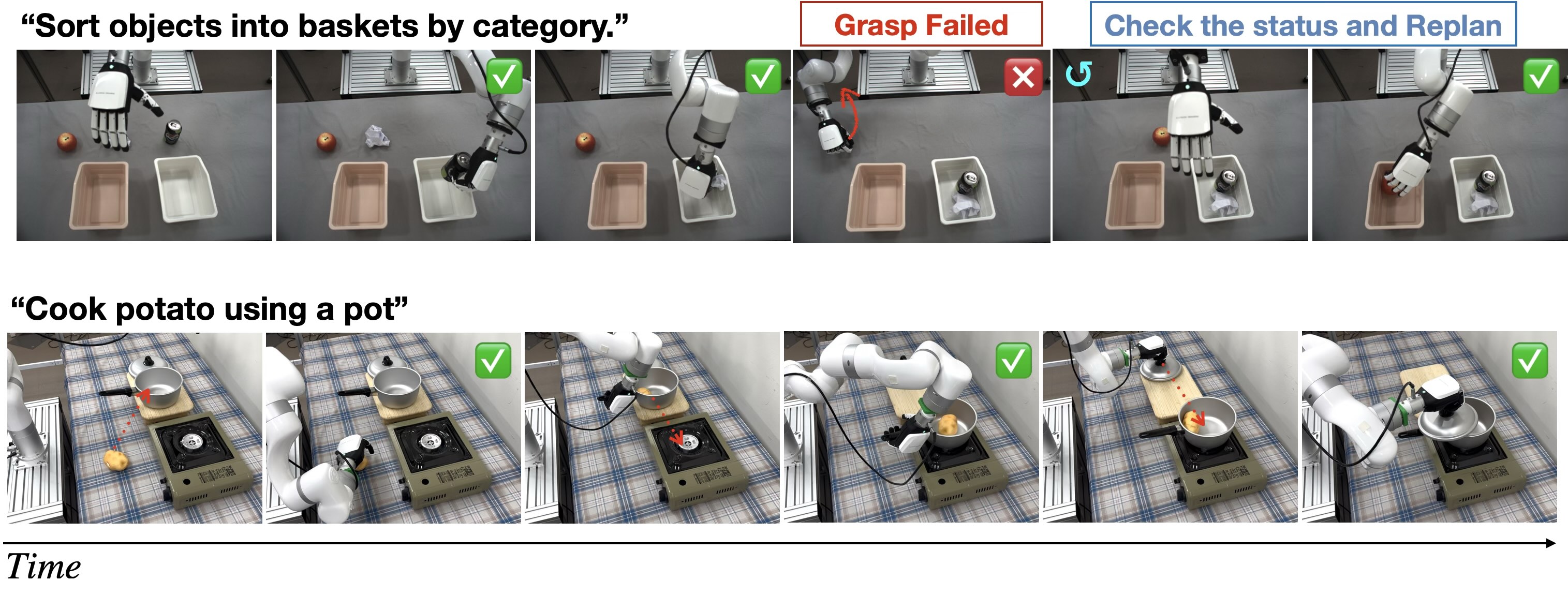}
\caption{\textbf{Qualitative Results.} Long-horizon manipulation examples. The shown scenarios consist of multiple subtasks. In the example above, the grasp fails, and the VLM detects the failure state and replans the next action.}
\label{fig:longterm_manip}
\end{figure}

We extend ZeroDex to long-horizon tasks through a planning and verification loop. The VLM $\Phi$ manages the remaining subtasks using current observations and a textual memory of prior attempts.
Given a high-level instruction $\mathbf{l}_{\mathrm{task}}$ and initial observations $\mathcal{I}_0$, $\Phi$ initializes an ordered subtask queue $\mathcal{L}^0$ and memory $h_0$. 
At attempt $k$, we select a pending subtask $\mathbf{l}^k$
from $\mathcal{L}^k$ and invoke
$(\mathcal{T}_{\mathrm{robot}}^k,\eta_k)
=\mathcal{Z}(\mathcal{I}_k,\mathbf{l}^k)$.
The generated trajectory is executed only if $\eta_k=1$.
After the attempt, we acquire new observations
$\mathcal{I}_{k+1}$, which allow $\Phi$ to assess the resulting
scene and update the memory and remaining plan:
\begin{equation}
    \begin{aligned}
        &(\alpha_k,h_{k+1},\mathcal{L}^{k+1}) =\Phi\bigl(
            \mathcal{I}_{k+1},\mathbf{l}_{\mathrm{task}},
            \mathcal{L}^k,h_k
        \bigr).
    \end{aligned}
    \label{eq:supp_closed_loop_update}
\end{equation}
The decision $\alpha_k$ selects $\it{continue}$, $\it{retry}$, $\it{replan}$, or $\it{done}$.
The updated queue guides subsequent calls to $\mathcal{Z}$, which grounds retried or revised subtasks again in the current scene. This loop supports recovery from intermediate failures, object displacement, and newly observed scene changes (\cref{fig:longterm_manip} and supplementary video).

\section{Experiments}

\begin{figure}
\centering
\includegraphics[width=1.0\linewidth]{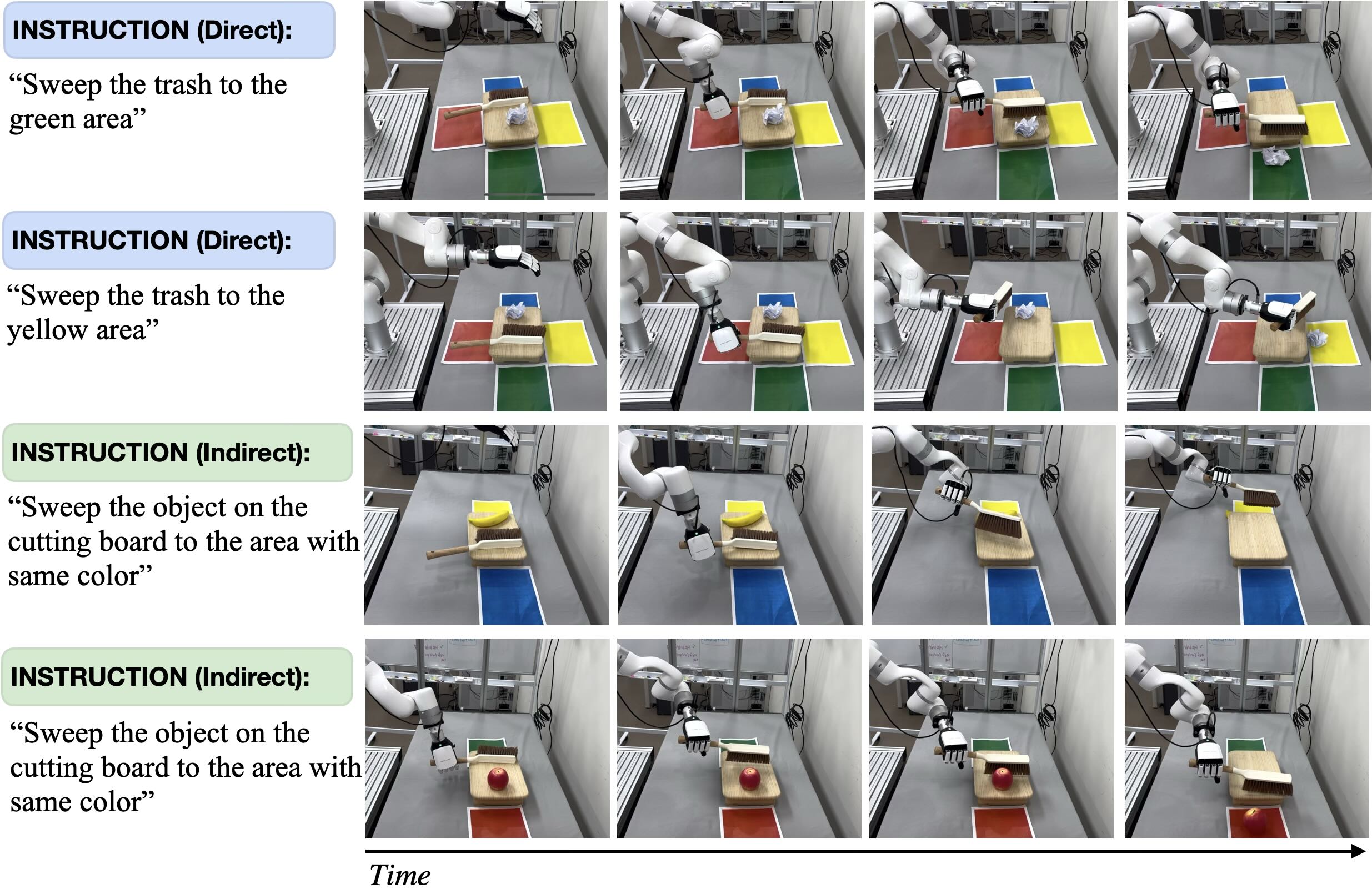}
\caption{\textbf{Qualitative Results.}  We evaluate both direct and indirect styles of instructions and lo successful 3D grounding across diverse environments.}
\label{fig:boaa_wide}
\end{figure}

We evaluate our framework on zero-shot robot manipulation in a real-world tabletop setting, assessing its scalability from simple tasks to long-horizon scenarios. Our evaluation covers four key capabilities: (1) target grounding amidst distractors and collision robustness (e.g., placing inferred object into a basket), (2) spatial-relation reasoning (e.g., placing a pot on a stove), (3) affordance-aware tool use (e.g., sweeping objects with a broom), and (4) long-horizon sequencing (e.g., cooking and organizing objects).

\subsection{Implementation Details}
Our framework is robot-agnostic apart from the final arm--hand motion generation, and we use two platforms: a Franka arm and an xArm, each equipped with an Inspire dexterous hand. All comparisons against zero-shot grounding baselines (\cref{tab:main_merged}, \cref{tab:failure_stage}) are conducted on the Franka. The comparison against fine-tuned VLA policies is conducted on the xArm, because our teleoperation rig for demonstration collection is built on that arm; our method and both VLA baselines are evaluated on the same xArm platform, scene list, and success criterion, so the comparison remains matched. Qualitative results use both arms. The tabletop environment is observed by four calibrated RGB cameras, including a stereo pair, and all experiments use this four-view configuration. We use FoundationStereo~\cite{wen2025foundationstereo} for stereo depth estimation and FoundationPose~\cite{wen2024foundationpose} for multi-object 6D pose estimation. All VLM queries, including reference-frame grounding, view-wise grounding, affordance prediction, and execution verification, use Gemini Robotics-ER~1.6 with medium thinking level. Arm--hand motions are produced by an arm--hand motion generator~\cite{curobo_v2,park2026natural}. The VLM prompts are provided in the supplementary video.

\subsection{Baselines}
We compare our zero-shot framework against three families of baselines.
(i)~\textbf{MOKA}~\cite{liu2403moka}. A zero-shot visual-prompting method whose keypoint-and-motion-primitive representation most closely matches ours. We follow its formulation, deprojecting object-surface keypoints from a single-view RGB-D depth map and letting the VLM select the height of the free-space waypoints. MOKA uses two options here (the same height as the target point, or above it) and leaves the actual offset unspecified; we instead provide a small set of discrete offsets so that taller objects and container rims can be cleared, which favors the baseline. Its grasp keypoint is expanded into a fixed-radius contact region and fed to the same affordance-based grasp generator, so the two methods differ only in grounding.
(ii)~\textbf{ReKep}~\cite{huang2024rekep}. A zero-shot method that expresses a task as relational constraints over keypoints deprojected from the same single RGB-D view, and solves for end-effector poses that satisfy them. We use the authors' released implementation for keypoint proposal and constraint generation, and route the resulting target poses through the same grasp and arm--hand execution stack as the other methods, so that ReKep and ours differ only in how the task is grounded in 3D. For other Zero-Shot Method, We exclude baselines that cannot be placed on the same execution stack or that require additional policy training.
(iii)~\textbf{VLA models.} GR00T~\cite{bjorck2025gr00t} and Being-H0~\cite{luo2025being}. We collect 100 teleoperated pick-and-place demonstrations (about one hour of trajectories) in which the target object identity is fixed while its position, the surrounding objects, and the placement vary. We post-train both models on this same set using their official configurations ($H{=}16$, batch 64, LR $10^{-4}$, 30k steps, $4\times$A6000; relative end-effector arm actions, 6-DoF absolute hand actions, two input views for GR00T, one for Being-H0, synchronous inference at 30\,Hz). Our method uses no robot demonstrations or weight updates.

\subsection{Metrics}
\label{sec:metrics}

\textbf{Success Rate.}
A trial is successful if the outcome required by the language instruction is achieved. 

\textbf{Grounding Quality.}
We assess groundings along two axes: reproducibility and physical validity. For reproducibility, we repeat the pipeline on identical images and compute the RMS deviation $\sigma$ of the lifted 3D grasp keypoint about its centroid, reporting $\sigma$ averaged over scenes together with the worst value $\delta_{\max}$ (mm). For physical validity, we pose the manipulated object at each predicted (unrefined) placement or tool waypoints and evaluate the environment penetration depth $\phi_m$ of \cref{eq:position_refinement}. For each scene, we compute the worst-case ($\mathcal{P}^o_{\max}$) and mean ($\mathcal{P}^o_{\mathrm{avg}}$) penetration depth (mm) over its waypoints, together with the collision rate Coll\%, the fraction of those waypoints exceeding $1$\,mm penetration. Table entries report these per-scene values averaged over scenes.

\textbf{Grasp Feasibility.}
For grasp candidates generated from a grounding, IK\% is the fraction for which collision-aware inverse kinematics succeeds both at the grasp pose and at the grounded waypoint with the object attached, and the clear rate is the fraction that is reachable and collision-free, i.e., with hand--scene penetration below the 5\,mm tolerance of the sphere-based hand collision model used by the motion generator.

\textbf{Long-Horizon Success Rate.}
For sequential tasks, a trial is considered successful only if all required steps are completed in the correct order. We report both the number of trials and the success rate. Unlike the single-step success rate, which is counted on the first attempt, we report long-horizon success both without retry and with a budget of two attempts per subtask, i.e., a single retry.

\subsection{Results}
\label{sec:result}

\begin{table}[t]
\centering
\footnotesize
\setlength{\tabcolsep}{4pt}
\caption{\textsc{Real-robot success rates.} $N$ is the number of task-specific teleoperation demonstrations for fine-tuning; for the fine-tuned VLA baselines we additionally report stage-wise success.}
\label{tab:main_merged}
\begin{tabular}{lcccc}
\hline
\textbf{Method} & $N$ & \textbf{Standard} & \textbf{Cluttered \& Precise} & \textbf{All} \\
 & & ($n{=}10$) & ($n{=}10$) & ($n{=}20$) \\
\hline
\multicolumn{5}{l}{\textit{Zero-shot grounding}} \\
MOKA~\cite{liu2403moka} & 0 & 0.60 & 0.20 & 0.40 \\
ReKep~\cite{huang2024rekep} & 0 & 0.50 & 0.60 & 0.55 \\
\textbf{Ours} & \textbf{0} & \textbf{0.90} & \textbf{0.80} & \textbf{0.85} \\
\hline
\multicolumn{2}{l}{\textit{Fine-tuned VLA}} & \multicolumn{3}{c}{\textbf{Throw Away Trash} ($n{=}20$)} \\
\cline{3-5}
 & & \textit{Approach} & \textit{Grasp} & \textit{Success} \\
GR00T~\cite{bjorck2025gr00t} & 100 & 0.95 & 0.35 & 0.30 \\
Being-H0~\cite{luo2025being} & 100 & 1.00 & 0.05 & 0.00 \\
\textbf{Ours} & 0 & 1.00 & 0.95 & 0.95 \\
\hline
\end{tabular}
\end{table}

\begin{table}[t]
\centering
\footnotesize
\setlength{\tabcolsep}{3pt}
\caption{\textsc{Where zero-shot methods fail.} Failure counts by stage; Success repeats Table~\ref{tab:main_merged}. Solver is ReKep's constraint-solving stage, which the others lack.}
\label{tab:failure_stage}
\begin{tabular}{lccccc}
\hline
\textbf{Method} & \textbf{Success} & \textbf{Ground} & \textbf{Solver} & \textbf{Grasp} & \textbf{Exec.} \\
 & ($n{=}20$) & & & & \\
\hline
MOKA~\cite{liu2403moka} & 0.40 & 6 & -- & 3 & 3 \\
ReKep~\cite{huang2024rekep} & 0.55 & 7 & 2 & 0 & 0 \\
\hline
\textbf{Ours} & \textbf{0.85} & \textbf{0} & -- & 3 & \textbf{0} \\
\hline
\end{tabular}
\end{table}

\begin{table}[!t]
\centering
\footnotesize
\setlength{\tabcolsep}{3pt}
\caption{\scriptsize\textsc{Effect of affordance conditioning.} Succ.\ is real-robot success over 10 trials. \#G: mean grasp candidates per scene; Clear\% and IK\% (Sec.~\ref{sec:metrics}) are pooled over all candidates in offline.}
\label{tab:affordance}
\begin{tabular}{lcccccccc}
\hline
& \multicolumn{4}{c}{\textbf{Tool-use}(5 scenes)} & \multicolumn{4}{c}{\textbf{Pick-and-place}(10 scenes)} \\
\cline{2-5}\cline{6-9}
\textbf{Affordance} & \#G & Clear\% & IK\% & \textbf{Succ.} & \#G & Clear\% & IK\% & \textbf{Succ.} \\
\hline
off & 22 & 16.7 & 16.7 & 2/10 & 69 & 23.4 & 5.6 & 9/10 \\
on & 16 & 96.2 & 35.0 & \textbf{7/10} & 27 & 96.0 & 37.2 & 9/10 \\
\hline
\end{tabular}
\end{table}

\begin{table}[t]
\centering
\footnotesize
\caption{3D grounding quality vs.\ MOKA~\cite{liu2403moka} on the same 15 scenes. All lengths in mm; metrics are defined in Sec.~\ref{sec:metrics}.}
\label{tab:grounding_quality}
\setlength{\tabcolsep}{2pt}
\resizebox{\columnwidth}{!}{%
{\renewcommand{\arraystretch}{1.27}%
\begin{tabular}[t]{@{}lcc@{}}
\toprule
\multicolumn{3}{@{}l}{\emph{Grasp repeatability}} \\
\multicolumn{3}{@{}l}{\scriptsize (10 scenes $\times$ 8 runs)} \\
\multicolumn{3}{@{}l}{\scriptsize } \\
Method & $\sigma\!\downarrow$ & $\delta_{\max}\!\downarrow$ \\
\midrule
MOKA~\cite{liu2403moka} & 8.1 & 64.9 \\
\textbf{Ours} & \textbf{2.5} & \textbf{3.9} \\
\bottomrule
\end{tabular}}%
\hspace{4pt}%
\begin{tabular}[t]{@{}lcccc@{}}
\toprule
Method & $\mathcal{P}^o_{\max}\!\downarrow$ & $\mathcal{P}^o_{\mathrm{avg}}\!\downarrow$ & Coll\%$\downarrow$ & IK\%$\uparrow$ \\
\midrule
\multicolumn{5}{@{}l}{\emph{Tool} (5 scenes)} \\
\quad MOKA~\cite{liu2403moka} & 20.4 & 20.4 & 20 & 30.3 \\
\quad \textbf{Ours} & 20.3 & \textbf{4.7} & \textbf{5} & \textbf{35.0} \\
\addlinespace
\multicolumn{5}{@{}l}{\emph{P\&P} (10 scenes)} \\
\quad MOKA~\cite{liu2403moka} & 37.2 & 36.3 & 100 & 24.2 \\
\quad \textbf{Ours} & \textbf{32.6} & \textbf{16.3} & \textbf{17} & \textbf{37.2} \\
\bottomrule
\end{tabular}}
\end{table}

\begin{table}[t]
\centering
\caption{Long-horizon success per sub-step and end-to-end. Main numbers allow one retry per subtask; numbers in parentheses are without retry.}
\label{tab:long_horizon_success}
\resizebox{0.9\columnwidth}{!}{%
\begin{tabular}{lcccc}
\toprule
Task & Step 1 & Step 2 & Step 3 & End-to-End \\
\midrule
Organize Objects & 10/10 & 10/10(7/10) & 8/10(6/7) & 8/10 (6/10) \\
Cooking (with tool-usage) & 9/10(8/10) & 9/9(7/8) & 9/9(6/7) & 9/10 (6/10) \\
\bottomrule
\end{tabular}}
\end{table}

\textbf{Real-robot success.}
Table~\ref{tab:main_merged} compares our method with the zero-shot grounding baselines on real-robot tasks. The standard condition uses uncluttered scenes with well-separated objects and no tall obstacles; the cluttered-and-precise condition adds surrounding distractors and raised, walled targets such as a basket. Our method reaches $0.85$ overall, against $0.40$ for MOKA and $0.55$ for ReKep, and leads on both conditions: $0.90$ on the standard condition, where MOKA and ReKep reach $0.60$ and $0.50$.
The advantage is most pronounced in cluttered scenes and tasks requiring precise placement, where multi-view 3D grounding localizes more reliably under occlusion and resolves the depth ambiguity of lifting 2D predictions to 3D: on the cluttered-and-precise condition our method achieves $0.80$, compared to $0.20$ for MOKA and $0.60$ for ReKep.
Table~\ref{tab:failure_stage} localizes the failures under the shared execution stack: six of MOKA's twelve failures and seven of ReKep's nine occur at grounding, whereas none of ours do. Grounding is the largest failure mode for both baselines, and more so for the stronger one: ReKep succeeds more often overall than MOKA yet attributes a larger share of its failures to grounding. Single-view depth lifting is brittle here: depth is discontinuous at object boundaries, so a few pixels of 2D error at the rim of a raised container drop the lifted point to the table behind it, placing the target in collision or out of reach. More fundamentally, lifting 2D groundings into 3D by deprojecting them and leaving the residual depth ambiguity to VLM reasoning is imperfect by nature.
Table~\ref{tab:main_merged} further shows that both VLA baselines, fine-tuned on 100 task-specific demonstrations, remain far behind: Being-H0 fails in all trials and GR00T reaches $0.30$, against $0.95$ for our zero-shot system. The stage-wise breakdown locates the gap at contact rather than perception: both policies approach the target in nearly every trial ($0.95$ and $1.00$) but secure a stable grasp in only $0.35$ and $0.05$ of them, underscoring how slowly VLAs adapt to a new dexterous embodiment and the value of a demonstration-free approach.

\textbf{3D grounding quality.}
Table~\ref{tab:grounding_quality} evaluates grounding repeatability and physical validity against MOKA.\footnote{ReKep is excluded from these metrics: its target comes from its constraint solver and is often specified as a range (e.g., 5--10\,cm above the keypoint), so these metrics would score clearance rather than grounding; see Tables~\ref{tab:main_merged} and~\ref{tab:failure_stage}.} Our multi-view fusion is about $3\times$ more repeatable than MOKA's single-view RGB-D lifting ($\sigma$ of $2.5$ vs.\ $8.1$\,mm; worst case $3.9$ vs.\ $64.9$\,mm) and yields lower object penetration and collision rates in both tool-use and pick-and-place (Coll\% of $5$ vs.\ $20$ and $17$ vs.\ $100$; $\mathcal{P}^o_{\mathrm{avg}}$ of $4.7$ vs.\ $20.4$ and $16.3$ vs.\ $36.3$\,mm). It also leaves more executable grasp candidates under the same collision-aware IK check: IK\% rises from $30.3$ to $35.0$ on tool-use and from $24.2$ to $37.2$ on pick-and-place. Collision-aware refinement (\cref{eq:position_refinement}) further raises motion-planning success from $7/10$ to $10/10$ over ten scenes by rescuing placements on raised surfaces (e.g., a cutting board), where the unrefined grounding puts the object inside the board instead of on top of it.

\textbf{Affordance conditioning.}
Table~\ref{tab:affordance} ablates the multi-view affordance region of \cref{sec:method:grasp}. Conditioning grasp generation on the affordance region raises the clear rate from $16.7\%$ to $96.2\%$ (tool-use) and from $23.4\%$ to $96.0\%$ (pick-and-place), and IK feasibility from $16.7\%$ to $35.0\%$ and from $5.6\%$ to $37.2\%$, respectively. On the tool-use tasks this translates into task success ($2/10$ to $7/10$), whereas pick-and-place is already saturated ($9/10$ in both settings). Multi-view grounding and affordance conditioning thus act on two independent axes: the former determines where objects are placed, while the latter determines whether the resulting grasps are reachable and collision-free.

\textbf{Long-horizon manipulation.}
Table~\ref{tab:long_horizon_success} reports per-step and end-to-end success rates for long-horizon tasks. With a single retry per subtask, end-to-end success is $8/10$ on Organize Objects and $9/10$ on Cooking, against $6/10$ for both without retry. Task Grounding errors and transient grasp failures are recovered by execution verification followed by re-grounding, retry, or replanning from the plan memory (\cref{sec:method:retry}), whereas arm joint limits and environment collisions remain the main unrecovered failures.

\textbf{Runtime.}
Multi-view grounding takes $19.4$ s per pass: $6.7$ s for reference-view grounding, $9.6$ s for the four per-view queries, and $3.1$ s for triangulation and the associated overhead. Arm--hand motion generation then spends $4.8$ s on scene construction and planner initialization, followed by $9.4$ s of planning that covers grasp batch IK, collision checking, collision-aware endpoint refinement, and trajectory computation. For dexterous grasp generation, we validate every candidate in physics simulation before execution to select more reliable grasps, so its cost grows with the number of candidates requested rather than with scene complexity, and we therefore do not report it as a single number.

\textbf{Extension to Non-Prehensile Manipulation.}
The same pipeline also covers non-prehensile behaviors such as pressing a button or pushing a drawer, which involve no grasp: we convert the grounded 3D primitive into a hand trajectory rather than an object trajectory. The corresponding atomic action stores a short sequence of hand poses instead of object poses and is aligned to the lifted primitives exactly as in the tool-use case, while execution bypasses grasp generation entirely and sends the aligned hand trajectory directly to the arm--hand motion generator. Qualitative examples are provided in the supplementary video.

\section{Discussion}
\label{sec:limitation}

We present ZeroDex, a zero-shot long-horizon manipulation framework that bridges VLM reasoning with physical execution via multi-view 3D grounding. By decomposing language instructions into sequences of 3D-grounded manipulation primitives, our system seamlessly supports both standard pick-and-place and complex tool-use tasks by spatially aligning object-centric atomic actions to the target scene. Experimental results demonstrate that our multi-view fusion strategy outperforms visual-prompting and constraint-based baselines in grounding repeatability, spatial accuracy, and robustness against occlusion, and that our demonstration-free system exceeds VLA policies fine-tuned on task demonstrations. Furthermore, the primitive-level formulation naturally supports execution verification with re-grounding and replanning from an explicit plan memory, allowing the system to verify task progress and recover from intermediate failures during long-horizon tasks.

\noindent \textbf{Limitations and Future Work.} Despite its capabilities, our framework exhibits a few limitations. First, the system's performance is inherently bounded by the reasoning reliability of the underlying 2D VLM. Although multi-view lifting and local collision-aware refinement improve the geometric consistency of grounded predictions, errors in task decomposition, affordance selection, or semantic grounding in 2D can still propagate to downstream execution failures. Second, the physical execution relies on external arm--hand motion generators, so kinematic singularities, collision-checker timeouts, or unstable grasp executions can still induce failures. Third, tool-use motions that are absent from the Bag of Atomic Actions cannot be executed directly. The library is, however, an expandable cache rather than a fixed vocabulary: new skills are added by recording object-centric trajectories from human demonstrations or by synthesizing them with VLMPose~\cite{vlmpose}, and the same mechanism supplies the hand-level primitives used for non-prehensile behaviors, whereas deformable objects, whose motion is not representable as a rigid 6D trajectory, remain future work. Fourth, our current formulation focuses on object-centric manipulation and tool use, and does not explicitly support dexterous in-hand manipulation, such as rotating an object within the hand or operating a held tool such as scissors. Future work will explore integrating native 3D vision-language models and reactive low-level policies to further enhance the speed and robustness of zero-shot dexterous manipulation.

\bibliographystyle{IEEEtran}
\bibliography{references,references_added}  %

\end{document}